\documentclass[journal]{IEEEtran}

\usepackage{graphicx}
\usepackage{amsmath,amsthm,amssymb}
\usepackage{multirow}
\usepackage{color}
\usepackage{subfig}
\usepackage{cite}
\begin{document}
\title{A Robust Approach for Securing Audio Classification Against Adversarial Attacks}

\author{Mohammad Esmaeilpour,
   Patrick Cardinal,
   and~Alessandro Lameiras Koerich,~\IEEEmembership{Member,~IEEE}
\thanks{M. Esmaeilpour, P. Cardinal and A. L. Koerich are with the Department of Software and IT Engineering, \'{E}cole de Technologie Sup\'{e}rieure (\'{E}TS), University of Quebec, Montreal, QC, Canada,
e-mail: (mohammad.esmaeilpour.1@etsmtl.ca, patrick.cardinal@etsmtl.ca, alessandro.koerich@etsmtl.ca).}
\thanks{Accepted for publication November, 2019.}}

\markboth{IEEE Transactions on Information Forensics and Security,~Vol.~X, No.~X, November~2019}%
{Esmaeilpour \MakeLowercase{\textit{et al.}}: IEEE Transactions on Information Forensics and Security}



\maketitle

\begin{abstract}
Adversarial audio attacks can be considered as a small perturbation unperceptive to human ears that is intentionally added to an audio signal and causes a machine learning model to make mistakes. This poses a security concern about the safety of machine learning models since the adversarial attacks can fool such models toward the wrong predictions. In this paper we first review some strong adversarial attacks that may affect both audio signals and their 2D representations and evaluate the resiliency of deep learning models and support vector machines (SVM) trained on 2D audio representations such as short time Fourier transform, discrete wavelet transform (DWT) and cross recurrent plot against several state-of-the-art adversarial attacks. Next, we propose a novel approach based on pre-processed DWT representation of audio signals and SVM to secure audio systems against adversarial attacks. The proposed architecture has several preprocessing modules for generating and enhancing spectrograms including dimension reduction and smoothing. We extract features from small patches of the spectrograms using the speeded up robust feature (SURF) algorithm which are further used to transform into cluster distance distribution using the K-Means++ algorithm. Finally, SURF-generated vectors are encoded by this codebook and the resulting codewords are used for training a SVM. All these steps yield to a novel approach for audio classification that provides a good tradeoff between accuracy and resilience. Experimental results on three environmental sound datasets show the competitive performance of the proposed approach compared to the deep neural networks both in terms of accuracy and robustness against strong adversarial attacks.
\end{abstract}


\IEEEpeerreviewmaketitle

\section{Introduction}
\IEEEPARstart{A}{dversarial} attacks pose security issues since they can be unrecognizable to human eyes or human ears while they can easily fool any trained machine learning model with very high confidence. As these machine learning models are becoming more present in many devices and applications, there exists an urgent need for improving their robustness against adversarial attacks. Basically, an adversarial attack algorithm formulates an optimization problem such as finding the smallest possible perturbation to be added to a given legitimate input (image, audio, spectrogram, etc.) aiming at a machine learning model to predict a wrong label. This perturbation should be as small as possible to be imperceptible to human visual or auditory system. Adversarial attacks have been attracting the attention of many researchers, mainly in the domain of computer vision \cite{kurakin2016adversarial,sabour2015adversarial, xie2017adversarial}. However, adversarial attacks may also pose a serious threat to voice assistant devices, speech and speaker recognition as well as other audio-related applications. In spite of that, few studies have addressed adversarial attacks for audio signals \cite{carlini2018audio}. One of the possible reasons is the considerable optimization overhead of adversarial algorithms when applied to audio signals due to their high dimensionality. In the big picture, adversarial examples of audio signals can be crafted during sound production or post production by changing their amplitude or frequency into the ranges where humans cannot perceive. This is difficult and needs to be treated carefully because there is no guarantee of producing a true adversarial example and the output could be just a noisy example. In the case of post-production of adversarial examples, the adversary can either solve an optimization problem (costly) or develop an adversarial filter in order to apply some perturbations to a legitimate audio before passing it through a machine learning model. In both cases, the victim model could be fooled toward the bad wishes of the adversary and make the system misbehave.

In this paper, we investigate the threat of adversarial attacks on environmental audio sounds due to the diversity that we may find, ranging from baby crying to engines, horns to dog barking or people chatting with numerical text-free labels. Adversarial attacks are quite useful for other relevant domains of speech recognition and music classification and they may be generalizable to speech-to-text applications, though the latter is not discussed in this paper. Environmental sound classification has been a challenging problem in machine learning research \cite{salamon2015unsupervised}. Both shallow and deep neural networks (DNNs) have shown competitive performances on benchmarking datasets such as ESC-10 \cite{piczak2015esc}, ESC-50 \cite{piczak2015esc}, and UrbanSound8K \cite{Salamon:UrbanSound:ACMMM:14}. Besides the supervised models, there are some unsupervised models such as spherical K-means for sound representation learning \cite{salamon2015unsupervised, salamon2015feature}. Both supervised and unsupervised models have mainly been trained either on audio waveforms (1D) or on 2D representation such as spectrograms. In both cases, convolutional neural networks (CNNs) have shown better performances compared to other classifiers. For instance, the CNN proposed by Salamon and Bello \cite{salamon2017deep} outperforms their prior approach based on unsupervised feature learning and random forest \cite{salamon2015unsupervised} on the UrbanSound8K dataset. Also, for ESC-10 and ESC-50 datasets, a 1D CNN with eight convolution layers (SoundNet) \cite{aytar2016soundnet} outperforms random forest \cite{piczak2015esc}, SVM using Mel-Frequency Cepstral Coefficients (MFCCs) \cite{piczak2015esc}, and convolutional autoencoders \cite{aytar2016soundnet}. In addition to these CNNs, other DNN architectures such as AlexNet and GoogLeNet, which have shown remarkable performances on image classification tasks (e.g. ImageNet dataset) have also been used for environmental sound classification. Interestingly, these two CNNs trained on spectrograms have been achieving the highest recognition performance for the three aforementioned datasets as reported by Boddapati \textit{et al.} \cite{boddapati2017classifying}.

One of the open problems in audio classification seemingly is no longer improving recognition accuracy but improving their strengths against some carefully crafted adversarial examples. Therefore, the proposed approach for environmental sound classification is based on two findings: (i) deep learning models, particularly AlexNet and GoogLeNet outperform conventional classifiers trained on handcrafted features such as SVM; (ii) SVM in general is more robust against adversarial attacks, potentially because it learns from low-dimensional feature vectors that might reduce the chance of being affected by adversarial perturbations compared to deep models which learn from raw data. Following these facts, in this paper we propose an SVM-based approach that provides a good tradeoff between the recognition accuracy and the robustness against adversarial attacks while achieving recognition accuracy comparable to deep models. Since there is no standard metric for evaluating the quality of such a tradeoff, we also introduce a distance metric based on the error rate versus the fooling rate.

Our contribution in this paper is threefold: (i) we present common adversarial attacks for audio and we show how they can affect the security of audio applications; (ii) we characterize the vulnerability of state-of-the-art models based on 2D representations to adversarial attacks and the transferability of these attacks between different models; (iii) we propose a novel approach for environmental sound classification that, in addition to being robust against several adversarial attacks without incorporating any reactive or proactive defense process, it also provides a high recognition accuracy, which is competitive with the state-of-the-art.

This paper is organized as follows. Section \ref{sec:aa} introduces general adversarial attacks and describe the most important ones. In this section we also present the adversarial attacks that may affect audio applications based on 2D audio representations and discuss adversarial attacks that may affect audio waveforms. Section \ref{sec:2daudio} presents the main 2D representations for audio signals. Section \ref{sec:robust} presents the proposed approach that aims of achieving both good classification accuracy and robustness to adversarial attacks. In Section \ref{sec:exp} we characterize the vulnerability of some state-of-the-art models in the problem of environmental sound classification, measure the resiliency of the proposed approach versus CNNs and review the adversarial example transferability among these models. The conclusions and perspectives of future work are presented in the last section.

\section{Adversarial Attacks}
\label{sec:aa}
Adversarial attacks can be considered as carefully crafted perturbations that when intentionally added to a legitimate example, lead machine learning models to misbehave \cite{Weng2018}. Considering $\bold{x}$ as a legitimate example, then an adversarial example $\bold{x}^\prime$ can be crafted in such a way that:

\begin{equation}
\bold{x} \approx \bold{x}^\prime, \qquad f^{*}(\bold{x})\neq f^{*}(\bold{x}^\prime)   
\end{equation}
\noindent where $f^{*}$ is the post-activation function. Supposing that $\bold{x}$ represents an image or an audio signal, the differences between $\bold{x}$ and $\bold{x}^\prime$ should not be perceived by the human visual or auditory systems. 

There are several algorithms for generating $\bold{x}^\prime$, mainly when $\bold{x}$ is an image. The adversarial attacks can be categorized into different groups. For instance, if the adversary has access to the model architecture, parameters, training dataset, etc., it is categorized as a white-box attack, otherwise it is called black-box. Also, adversarial attacks can have other taxonomy such as targeted, where the adversarial perturbation is crafted having in mind a specific target label, and non-targeted, where the adversarial perturbation is crafted to induce a machine learning model to predict any incorrect label. Due to the importance of studying adversarial threats for data-driven machine learning models, many attack algorithms have been proposed and they have shown a great success in fooling advanced models. However, the main challenge of almost all attack algorithms is their computational complexity, which makes adversarial training very time-consuming.

One of the first proposed attacks is the Fast Gradient Sign Method (FGSM) \cite{goodfellow2014explaining}, which still remains one of the most effective attacks. FGSM was originally built to attack CNNs but it can also be a serious threat for non-deep architectures. FGSM generates an adversarial example ${\bold{x}}^\prime$ by:

\begin{equation}
\bold{x}^\prime = \bold{x}+\epsilon \cdot \mathrm{sign}(\nabla_{\bold{x}}J(\bold{w}, \bold{x}, y))
\label{FGSM}
\end{equation}
\noindent where $\bold{x}$ and $y$ are the legitimate input and its true label respectively, $\epsilon$ is a constant value which can be determined by an optimization scheme, and $J$ is the cost function for the model parameter $\bold{w}$ obtained after completing the training process. FGSM is a white-box attack which means that the model parameter $\bold{w}$ should be accessible to fetch its gradient information and generate the adversarial example $\bold{x}^\prime$. In other words, by providing the trained model and the training dataset, FGSM can generate adversarial examples $\bold{x}^\prime$ using Eq.~\ref{FGSM}, which have unrecognizable differences to the legitimate input $\bold{x}$ and $\bold{x}^\prime$ can perhaps make the model $\bold{w}$ to predict a wrong label $y^\prime\neq y$ with high confidence.

The iterative version of the FGSM attack is known as Basic Iterative Method (BIM) \cite{kurakin2016adversarial} and its attack frequency is higher than one. In fact, BIM's optimization procedure can stop after generating the first adversarial example (BIM-a) or continue up to a pre-defined number of iterations (BIM-b). These two attacks are actually the improved version of FGSM which increases the attack rate to the cost of higher computational complexity.

Carlini and Wagner \cite{carlini2017towards} have proposed an optimization-based attack known as CWA, which uses the similarity metric $d_i$ defined in Eq.~\ref{CW_dif}.
\begin{equation}
d_{i}=\left \| \bold{x}_{i}-\bold{x}^\prime_{i} \right \|
\label{CW_dif}
\end{equation}
\noindent where $i$ is the sample index. CWA attempts to minimize $d_i$ as:
\begin{equation}
\underset{c}{\min} \left \| d_{i} \right \|+c \times g(\bold{x}+d_{i}) \quad \mathrm{s.t.} \quad  \bold{x}+d_{i} \in \left [ 0,1 \right ]^{n}
\label{CW_eq}
\end{equation}
\noindent where $c>0$ is a suitably chosen constant, $g(d) \geq 0 \iff f(d)=y^\prime$; and $y^\prime$ is the wrong label for $\bold{x}$. The intuition behind Eq.~\ref{CW_eq} is similar to the dropout variational inference introduced by Li \textit{et al.} \cite{li2017dropout}. This attack is very similar to the FGSM attack with two main differences: (i) it changes the input $\bold{x}_{i}$ using the $\tanh$ function; (ii) it uses a difference between logits (the vector of non-normalized predictions that a model generates) instead of optimizing a cost function for regular cross-entropy. CWA is one of the strongest iterative and targeted adversarial attacks and it can be very effective in fooling CNNs, though costly as it might need too many callbacks to $\bold{x}$. 

\begin{figure}[htpb!]
  \centering
  \includegraphics[width=0.25\textwidth]{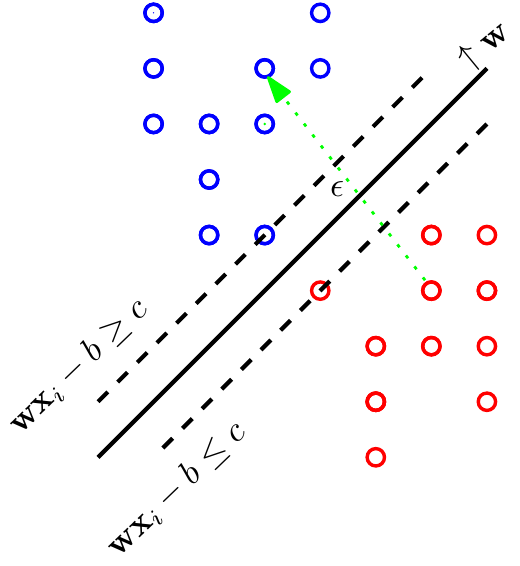}
  \caption{Simplified visualization of hard margin form of SVM adversarial attack. The dotted arrow depicts the Eq.~\ref{evasion_attack}. Datapoints are represented in two colors: red (adversarial) and blue (legitimate).}
  \label{SVM_model_att}
\end{figure}

The adversarial attacks presented so far are designed for DNNs. Since the approach proposed in this paper is based on SVMs, we also present two adversarial attacks designed to attack SVM models: Evasion attack (EA) and Label Flipping attack (LFA). EA \cite{biggio2013evasion} and LFA \cite{xiao2012adversarial}. The main difference between these two attacks is that LFA contaminates the training data by flipping the true labels of the samples, while EA manipulates the sample distribution aiming to change the true labels. In both cases, the decision boundary of the model is shifted toward maximum loss for the test set. The general intuition behind EA is to map an input $\bold{x}$ over a support vector(s) by simply flipping its label. This flipping can be toward the trained weight direction(s) of the SVM as given by Eq.~\ref{evasion_attack}.
\begin{equation}
\bold{x}^\prime = \bold{x} - \epsilon \odot  \frac{\bold{w}_i}{\left \| \bold{w}_i \right \|}
\label{evasion_attack}
\end{equation}
\noindent where $\bold{x}^\prime$ is the crafted adversarial example, $\bold{w}_{i}$ is the weight vector discriminating support vectors, and $\epsilon$ is a small constant value. The intuition behind these two attacks is the geometrical definition of support vectors as given by Eq.~\ref{SVM_opt_def}.
\begin{equation}
\min \bold{w}
\qquad \mathrm{s.t.} \quad y_{i}(\bold{w}^\top\bold{x}_{i} - b) \geq 1 \quad i=1,\dots, n 
\label{SVM_opt_def}
\end{equation}
\noindent where $\bold{w}$ is a vector normal to the hyperplane ($\bold{w}^\top\bold{x}-b=0$), $b$ is a bias term, and $y=\{+1,-1\}$ is the label. The position of the support vectors can be depicted as shown in Fig.~\ref{SVM_model_att}. 


In other words, the SVM model will be fooled by moving a datapoint perpendicularly toward the opposite direction of its weight vector. This attack is generalizable to soft margin SVM by simply optimizing the value of $\epsilon$ in Eq.~\ref{soft_Margin_svm}.
\begin{equation}
\frac{1}{n}\sum_{i=1}^{n}\max(1-y_{i}(\bold{w}^\top\bold{x}-b),0)+\epsilon\left \| \bold{w} \right \|^{2}
\label{soft_Margin_svm}
\end{equation}
As long as the optimization of $\epsilon$ is perpendicularly directed toward the $\bold{w}_i$, the SVM model cannot distinguish an adversarial from legitimate examples. This data contamination in EA can be implemented by taking advantage of gradient information and local search for achieving the best data perturbation with a specific budget as introduced by Biggio \textit{et al.} \cite{biggio2013evasion}. The gradient information can be exploited for different kernels. For an RBF kernel with variance $\sigma^{2}$, we have $K(\bold{x},\bold{x}_i)=\exp(-0.5 \cdot \sigma^{-2} \left \|\bold{x}-\bold{x}_i\right\|^{2})$, and its gradient can be computed by Eq.~\ref{RBF_grad}. 
\begin{equation}
\nabla K (\bold{x},\bold{x}_i)=-\sigma^{-2}\exp(-0.5 \cdot \sigma^{-2} \left\|\bold{x}-\bold{x}_i\right\|^{2})(\bold{x}-\bold{x}_i)
\label{RBF_grad}
\end{equation}
\noindent Similarly, for a polynomial kernel of degree $p$, denoted as $K(\bold{x},\bold{x}_i)$=$(\left\langle\bold{x},\bold{x}_i\right\rangle+c)^{p}$, its gradient can be computed by Eq.~\ref{nonlin_grad}.
\begin{equation}
\nabla K(\bold{x},\bold{x}_i)=p(\left\langle\bold{x},\bold{x}_i\right\rangle+c)^{p-1}\bold{x}_i
\label{nonlin_grad}
\end{equation}
\noindent Therefore, the adversarial example $\mathbf{x}^\prime$ can be computed by:
\begin{equation}
\bold{x}^\prime=\bold{x}-\eta \nabla f(\bold{x})
\end{equation}
\noindent where $\eta$ is a small scalar (step size) and $f$ denotes the learned filters in the hypothesis space $H$ for $K(\bold{x},\bold{x}_i)$=$\Phi(\bold{x})^\top \Phi(\bold{x}_i)$ and $\Phi$ is a mapping function from input to the feature space. Unlike EA, LFA does not generate an adversarial example via distorting the legitimate samples, but it contaminates the labels of such samples. This should result in maximum loss in the test set while it is expected to be minimum for the training set. The LFA attack can be implemented by solving the following optimization problem:
\begin{equation}
\min_{q,\bold{w},\epsilon,b} \gamma\sum_{i=1}^{2n}q_{i}(\epsilon_{i}-\xi_{i})+\frac{1}{2}\left \|\bold{w}\right\|^{2}
\end{equation}
\noindent subject to:
\begin{equation}
y_{i}(\bold{w}^\top\bold{x}_i+b)\geq 1-\epsilon_{i} \quad \epsilon_{i}\geq 0, \quad i=1, \cdots, 2n 
\end{equation}
\noindent having the budget of:
\begin{equation}
\sum_{i=n+1}^{2n}c_{i}q_{i} \leq C 
\end{equation}
\noindent where $\gamma$ is a fixed positive parameter for quantifying the trade off, $q_{i}$$\in$$\left\{0,1\right\}$ is an indicator variable for controlling over the legitimate ($q$=$0$) and contaminated example ($q$=$1$), and $c_{i}$ and $C$ denote the flipping cost of each example and the total flipping cost, respectively, from adversary's point of view. The hinge loss function ($\mathcal{L}$), defined in Eq.~\ref{hinge_loss} 
\begin{equation}
\mathcal{L}(y, f(\bold{x})) :=  \max(0, 1-y f_{D}(\bold{x}))
\label{hinge_loss}
\end{equation}
\noindent This loss function has been also used for $\epsilon_{i}$ on the contaminated dataset of $D^\prime$ as:
\begin{equation}
\epsilon_{i} :=  \max(0, 1-y_{i} f_{D^\prime}(\bold{x}_{i}))
\label{hinge_loss_sp}
\end{equation}
\noindent where ${D^\prime}$ is the contaminated dataset which also includes the original dataset $D$. Similarly, $\xi_{i}$ refers to the hinge loss of the classifier $f_{D}$: 
\begin{equation}
f_{D}(\bold{x}) := \bold{w}^{\top}\bold{x}+b  \text{,\ \ \ \ \ } \bold{w}:=\sum_{i=1}^{n}\alpha_{i}\Phi(\bold{x}_i).
\end{equation}
\noindent Herein, $b$ is the bias term and $\alpha$ denotes the Mercer kernel coefficient of the SVM.

\subsection{Transferability of Adversarial Attacks}
\label{subsec:transfer}
One of the main characteristics of adversarial attacks described so far is that they are non-targeted toward a specific label as they maximize the probability of any label other than the true one. This is very tricky since it opens up the opportunity of adversarial transferability to other data-driven models. This means that adversarial examples maintain their effectiveness against models different from those targeted by the attack. For instance, the FGSM attack, which targets CNNs, could completely fool a maxout network trained on the MNIST dataset \cite{goodfellow2014explaining}. Goodfellow \textit{et al.}~\cite{goodfellow2014explaining} have shown that the linear behavior of FGSM can be transferred to other classifiers including SVMs even with radial basis kernel function. This was a breakpoint of studying adversarial transferability for all classifiers, from logistic regression (simple) to very-deep CNNs (complex). Recently, Sabour \textit{et al.} \cite{sabour2015adversarial} have shown the great effectiveness of FGSM on fooling other deep architectures with and without convolution layers. 

A lot of effort has been made on improving transferability of adversarial attacks. From expanding input patterns (data-wise) \cite{xie2018improving} to developing ensemble models that produce more misleading adversarial examples (model-wise) \cite{liu2016delving}. Therefore, this is a real threat since adversarial attacks can be transferred among almost all models, e.g. from CNN to SVM, logistic regression and decision trees \cite{papernot2016transferability}. Besides that, models trained for speech-to-text translation have also been successfully fooled by crafted adversarial examples \cite{carlini2018audio}. Empirically, machine learning models designed for audio applications, based on either 1D or 2D representation are very vulnerable against adversarial attacks and the current defense schemes, such as those proposed by Das \textit{et al.} \cite{das2018adagio}, do not work appropriately. 

\subsection{Adversarial Attacks for Audio Signals}
\label{subsec:audioatt}
Adversarial attacks have been mainly studied in the domain of computer vision to perturb images. It has been shown that 2D CNNs are quite vulnerable against white-box and black-box optimization-based attacks \cite{goodfellow2014explaining}. However, these optimization-based attacks are usually very costly, and they require too many callbacks to each legitimate example, pixel-by-pixel. Generalizing these optimization-based attacks to audio signals (1D) is not straightforward since the audio signal is usually high-dimensional data, even considering a single audio channel. For instance, five seconds of mid-quality audio corresponds to an array of 110,250 points. Therefore, computing a similarity measure such as the $\ell_2$-norm between legitimate and crafted examples as a part of an adversarial optimization criterion is very challenging compared to 2D arrays. 

Alzantot \textit{et al.} \cite{alzantot2018did} and Du \textit{et al.} \cite{du2019sirenattack} have proposed speech-to-text adversarial attacks where the optimization process is replaced with heuristic algorithms like genetic algorithms \cite{alzantot2018did} or particle swarm optimization \cite{du2019sirenattack} to mitigate the considerable cost of the optimization process. Basically, these greedy and evolutionary algorithms introduce random noise to a legitimate example which in turn increases the chance of having a dissimilarity between legitimate and crafted adversarial examples. However, this also paves the way for an easy detection of adversarial examples by simple algorithms. On the other hand, in the most effective adversarial attacks for images (e.g. FGSM, BIM, CWA, etc.), adversarial perturbations are generated by an optimization process that has two key constraints: (i) induce a machine learning model to produce a wrong label; (ii) have a visual similarity between legitimate and adversarial examples. 

It is difficult to satisfy these constraints for adversarial audio because it is very challenging and time-consuming optimizing for these two constraints considering the high dimensionality of audio signals. Moreover, in contrast with images, audio signals are not convolved in rows and columns and this also makes very difficult solving the optimization problem for adversarial audio perturbations. These difficulties constitute enough ground for introducing evolutionary algorithms to randomly search for possible adversarial perturbations which basically can only respect the first key constraint. The main side effect of this approach is producing adversarial examples that stay close to the manifold of legitimate samples that can be easily detected by a tuned classifier or by a simple adversarial detector such as downsampling or upsampling. In this case, adversarial examples crafted by greedy algorithms lie in the submanifolds close to the legitimate samples, which is basically the same manifolds where noisy samples lie in. 

Some adversarial attacks explicitly add noise to the audio signals mainly by manipulating the frequency components \cite{roy2017backdoor,song2017inaudible}. Backdoor attack \cite{roy2017backdoor} is based on adding non-linearity to an audio signal in frequency ranges inaudible to the human auditory system (over 20 kHz). This non-linearity can be captured by microphones but does not show recognizable effects on human ends. Taking advantage of this type of attack, perturbations can be computed in frequency domain and then applied to an audio signal, which can fool a machine learning model. Backdoor attack lacks in defining a general optimization formulation for computing adversarial frequency perturbations (the shadow signal) \cite{roy2017backdoor}. In other words, there is no analytical way for computing the perturbation. The potential perturbation value may change depending on the audio signal and therefore it makes the computation of proper shadow signals very cumbersome and time-consuming. Moreover, audio frequency manipulation, even if unrecognizable by humans, can be easily detected if the perturbed audio signal is converted to a 2D representation. For instance, adversarial examples generated by the Backdoor attack can be easily detected by a simple post-processing module which analyzes their spectrograms. An ideal case for an adversarial audio example is to be unrecognizable in both 1D and 2D representations. Similarly, DolphinAttack \cite{song2017inaudible} implements phase domain manipulations to change the sample label toward other than the legitimate one that is unrecognizable by the human auditory system.

The detectability of the adversarial examples generated by Backdoor and DolphinAttack algorithms can be assessed by computing the local intrinsic dimensionality score (LID) \cite{ma2018characterizing} for their 2D representations. For such an aim, three groups of inputs should be defined: normal, noisy and adversarial where the latter is generated by both Backdoor and DolphinAttack algorithms. Next, each sample can be divided into mini-batches and the LID score can be computed for each mini-batch of these three groups with respect to their corresponding legitimate examples, by Eq.~\ref{LID}: 

\begin{equation}
\mathrm{{LID}}(\bold{x})=-{\left(
\frac{1}{k}\sum_{i=1}^{k}\log \frac{r_{i}(\bold{x})}{r_{k}(\bold{x})}\right)}^{-1}
\label{LID} 
\end{equation}
\noindent where $\bold{x} \in\Re^{n\times m}$ is a 2D array, $r_{i}(\bold{x})$ refers to the distance between $\bold{x}$ and their nearest neighbors, $r_{k}(\bold{x})$ denotes the maximum of the neighbor distances, and $k$ is the number of neighbouring samples. The LID scores of noisy and normal samples should be appended into negative class; and the LID scores of adversarial samples should be assigned to the positive class. Finally, a logistic regression can be trained on these two classes. The experiments carried out on 2D representations of audio signals in Section \ref{sec:detaudio} show that the adversarial examples generated by Backdoor and DolphinAttack can change the true label, although they cannot be categorized as adversarial attacks because of two main reasons: (i) the adversarial examples lie in the subspace of legitimate and random noisy signals when they basically should lie into different sub-regions; (ii) since there is not an analytical or an optimization-based approach for computing small adversarial perturbations for high-dimensional audio, the values of such perturbations are actually generated manually or by greedy algorithms and therefore, this highly increases the chance of detecting the adversarial signal even by a simple defense model. 

As it has been discussed so far, there are many open problems in crafting adversarial perturbations to raw audio signals and there is no reliable adversarial attack to 1D signals. This could also be interpreted as a good point if we disregard the fact that audio can be converted to a 2D representation (spectrogram) where strong adversarial attacks developed for images (e.g. FGSM, BIM, etc.) are quite applicable for 2D audio representations. This is a critical issue and poses a security concern for machine learning models for audio, either shallow (e.g. SVM) or deep learning models (e.g. CNNs). However, addressing the transferability of adversarial examples from 1D audio signals to 2D audio representations (or vice versa) is out of the scope of this paper. In fact, one of our goals in this paper is to assess the resiliency of machine learning models based on different types of 2D audio representations to some strong adversarial attacks aiming to better understand their vulnerabilities. 

\section{2D Audio Representation}
\label{sec:2daudio}
The vulnerability of machine learning models such as CNNs and long short-term memory networks on audio waveforms has been studied by Carlini and Wagner \cite{carlini2018audio}. They have shown the weaknesses of these models against FGSM-like adversarial attacks. However, the state-of-the-art for several audio tasks, such as music genre classification \cite{liu2019bottom,costa2012}, speaker identification \cite{sengupta2019speaker}, environmental sound classification \cite{sengupta2019speaker}, etc. are based on 2D representation. This aroused our interest to evaluate the robustness of models based on 2D representations against adversarial attacks. To the best of our knowledge, the resiliency of 2D CNNs such as AlexNet and GoogLeNet, which have achieved the highest performances on environmental sound datasets, against adversarial attacks has not been studied in 2D representation spaces. To such an aim, we use Fourier and wavelet transforms to convert raw audio signals into 2D representations. The first transformation is used to produce short-time frequency spectrograms for training AlexNet and GoogLeNet \cite{boddapati2017classifying}. We also use wavelet transform for producing more informative spectrograms, which after some pre-processing steps are used in the proposed approach to train an SVM classifier. A brief description of these two types of spectrogram is presented as follows. 

Considering a discrete-time audio signal $a[n]$, where $n = 0,1,\dots,N - 1$ denotes the number of samples and its decomposed signal $S$ using Fourier (time-frequency) transform using $\left \{ g_{\tau,\varrho} \right \}_{\tau,\varrho}$ atoms, as:
\begin{equation}
S[\tau,\varrho] = \left \langle a,g_{\tau,\varrho} \right \rangle = \sum_{n=0}^{N-1}a[n]g_{\tau,\varrho}^{*}[n]
\label{furier_transform}
\end{equation}
\noindent where the operator $*$ denotes the complex conjugate, and $\tau,\varrho$ are time and frequency localization indices, respectively. This representation is widely used in sound and speech processing \cite{yu2008audio2, mallat2008wavelet}. Given a Hanning window $H[n]$ of size $\vartheta$ which is shifted by a step $u \leq \vartheta$, then $\left \{ g_{\tau,\varrho} \right \}_{\tau,\varrho}$ in the latter equation can be defined as \cite{yu2008audio}:
\begin{equation}
g_{\tau,\varrho}[n] = H[n-\tau u]\exp\left ( \frac{j2\pi \varrho n}{\vartheta} \right )
\label{mother}
\end{equation}
\noindent where $0 \leq \tau \leq N/u$ and $0 \leq \varrho \leq \vartheta$ denote bindings of time and frequency (scale) indices respectively. Finally, the Fourier spectrogram is represented as:
\begin{equation}
{\bold{sp}}_\text{STFT}[\tau,\varrho]=\log \left | S[\tau,\varrho] \right |
\label{spec_furier}
\end{equation}
The final appearance of a spectrogram depends on the parameters $\tau$ and $\varrho$. Similar to this transform is the continuous wavelet transform ($CWT$) as denoted in Eq.~\ref{spec_wavelet}: 
\begin{equation}
CWT(\mho ,z; a(t),\psi(t))=\frac{1}{\sqrt{\mho }}\int_{-\infty}^{+ \infty}a(t)\psi(\frac{t-z}{\mho })dt
\label{spec_wavelet}
\end{equation}
\noindent where $\psi(t)$ denotes the mother wavelet and $\mho$, $z$, and $t$ stand for scale, translation and time, respectively. The discretized representation of $CWT$ is given by Eq.~\ref{DWT}, and it is determined on a grid of $\mho$ scales and $n$ discrete time with dilation parameter $\rho$.
\begin{equation}
DWT(\mho,n)=2^{\mho/2}\sum_{\rho = 0}^{n-1}a(\rho)\psi(2^{\mho},\rho -n)
\label{DWT}
\end{equation}
For $\psi$, we use Morlet function where $\mho$ is set to 0.8431:
\begin{equation}
\psi(t) = e^{-(\mho^{2}t^{2})/2}\cos (j \pi t)
\label{morlet}
\end{equation}
Finally, the wavelet spectrogram can be obtained as:
\begin{equation}
{\bold{sp}}_\text{DWT}[\mho,z] = \left | DWT(\mho,z) \right |^{2}
\label{wave_spp}
\end{equation}
In summary, for an audio signal $a[n]$, there will be two different 2D representations: $\bold{sp}_\text{STFT}$ and $\bold{sp}_\text{DWT}$. Moreover, for the latter spectrogram we use three scales for the magnitude, which provide different visualization schemes: linear, logarithmic, and logarithmic real. Linear scale highlights high-frequency magnitudes which represent high variation areas in the spectrogram. Logarithm scale highlights low-frequency information which expands distance of magnitudes in different scales. Finally, logarithm real scale highlights the energy of the signal which is related to the signal’s mean.

\section{A Robust Approach for 2D Audio Representation and Classification}
\label{sec:robust}
\begin{figure*}[htpb!]
  \centering
  \includegraphics[width=0.9\textwidth]{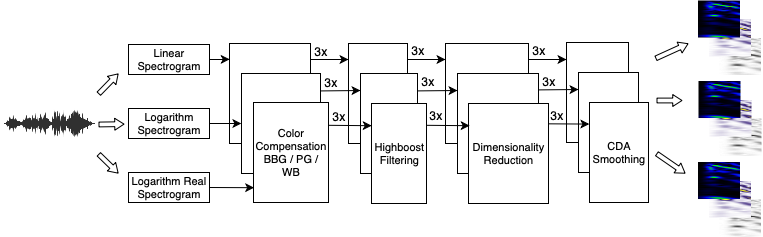}
  \caption{Overview of spectrogram generation and preprocessing. From a single audio waveform, three spectrogram representations are generated and processed through several blocks with the aim of enhancing the 2D representation.} 
  \label{PreprocDiag}
\end{figure*}

\begin{figure*}[htpb!]
  \centering
  \includegraphics[width=0.8\textwidth]{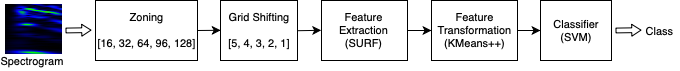}
  \caption{Overview of the proposed classification approach. Values in the first block indicate sizes of square zones (blocks) from 16$\times$16 to 128$\times$128. Stride values in the second block correspond to the zone sizes in the first block. For instance, a 96$\times$96 block has stride 2, and so on.} 
  \label{Flowchart_Classifier}
\end{figure*}

In general, the current approaches for audio classification are able to achieve high accuracy but they are vulnerable to adversarial attacks, what means that they can be easily fooled by adversarial examples. Therefore, our aim is to design a novel approach for audio classification that provides a good tradeoff between classification accuracy and low vulnerability to some of the most threatening adversarial attacks. The proposed approach for environmental sound classification has three main parts: spectrogram preprocessing, feature extraction, and classification. Fig.~\ref{PreprocDiag} presents an overview of the proposed preprocessing approach which, given an audio signal produces three spectrogram representations as output. The audio signal undergoes through color compensation, highboost filtering, dimensionality reduction, and smoothing and at the end, we have three enhanced spectrograms. Next, speeded up robust features (SURF) are extracted from zoning blocks that slide over the spectrograms as shown in Fig.~\ref{Flowchart_Classifier}. The geometrical distance of feature vectors is maximized by a K-means++ algorithm and finally a multiclass SVM trained on such features makes the prediction.  

\subsection{Spectrogram Preprocessing}
\begin{figure}[htpb!]%
    \centering
    \subfloat[]{{\includegraphics[width=0.115\textwidth]{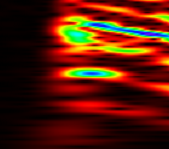} }}%
    \subfloat[]{{\includegraphics[width=0.115\textwidth]{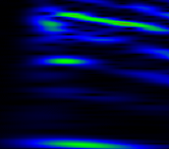} }}%
    \subfloat[]{{\includegraphics[width=0.115\textwidth]{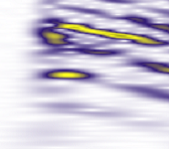} }}%
    \subfloat[]{{\includegraphics[width=0.115\textwidth]{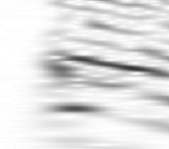} }}%
\caption{Spectrogram examples: (a) original; (b) black-blue-green (BBG); (c) purple-gold (PG); (d) white-black (WB).} 
\label{color_space}
\end{figure}

The goal of the spectrogram preprocessing is threefold: (i) improve the accuracy of the front-end classifier; (ii) improve the robustness of the trained model against adversarial attacks; (iii) artificially increase the number of samples of the dataset. It starts by color compensation of the spectrogram $\bold{sp}_\text{DWT}$ by mapping each spectrogram to three different color spaces: black-blue-green (BBG), purple-gold (PG), and white-black (WB) as shown in Fig.~\ref{color_space}. Empirically, color compensation boosts and improves the final classification performance because it affects pixel intensity values, though keeping their distributions. The second preprocessing operation is highboost filtering \cite{gonzalez2016digital}, which enhances color compensated spectrograms focusing on their high-frequency elements while maintaining low-frequency components. The output of the filter is denoted as ${\bold{sp}}_\text{ENH}$ which is given by Eq.~\ref{highboost}. 

\begin{equation}
{\bold{sp}}_\text{ENH} = (F_{ap} + cF_{hf})\times {\bold{sp}}_\text{DWT}
\label{highboost}
\end{equation}

\noindent where $F_{hf}$ represents a high-pass filter (5$\times$5 Laplacian operator) which is multiplied by a constant value $c$ which acts as a scaling factor, and $F_{ap}$ denotes an all-pass filter. 

\begin{figure}[htpb!]%
    \centering
    \subfloat[]{{\includegraphics[width=0.115\textwidth]{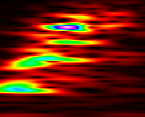} }}%
    \subfloat[]{{\includegraphics[width=0.115\textwidth]{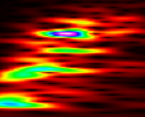} }}%
    \subfloat[]{{\includegraphics[width=0.115\textwidth]{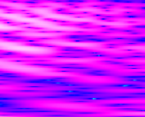} }}%
    \subfloat[]{{\includegraphics[width=0.115\textwidth]{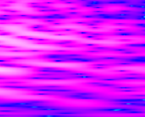} }}%
\caption{Dimension reduction effect: (a) linear magnitude representation; (b) reconstruction of (a) after reduction in half; (c) logarithmic magnitude representation; (d) reconstruction of (c) after reduction in half.} 
\label{color_space_recon}
\end{figure}

The three spectrogram representations and color compensations increase in nine times the number of samples into the datasets in addition to the pitch-shifting augmentation that is also applied, but on the 1D signal prior to the spectrogram representation. Pitch-shifting increases by eight times the number of samples. Therefore, to alleviate the computational complexity both in computing and storage, we reduce the dimensionality of the spectrograms. Though, there are many algorithms for such an aim, we use singular value decomposition (SVD) because of its pivotal properties in reducing the dimensionality of 2D data without changing the perceived visual appearance, if the reduction rank is chosen appropriately. Somewhat similar to the Fourier transform, SVD can describe a 2D matrix by basis functions in such a way that, linear combination of these functions can reconstruct the original spectrogram \cite{esmaeilpour2013new}. Basis functions in Fourier transform are sine and cosine, but SVD produces individual basis matrices for each given input. For an enhanced spectrogram $\bold{sp}_\text{ENH}$, SVD decomposes it as:
\begin{equation}
{\bold{sp}}_\text{ENH} = \sum_{i=1}^{m{}'}G_{i}U_{i}V_{i}^{\top}
\label{svd}
\end{equation}
\noindent where $G$, $U$, and $V$ are derived matrices from decomposing $\bold{sp}_\text{ENH}$ into singular value, hanger, and aligner matrices, respectively. Also $m{}'$ is the minimum dimension of the spectrogram either in width or height. The matrix $G$ is a diagonal and its elements are in descending order which indicates the importance of hanger and aligner column vectors. The basis functions associated with $\bold{sp}_\text{ENH}$ are the product of $U_{i}$ and $V_{i}^{\top}$ weighted by $G_{i}$. This allows us to reconstruct $\bold{sp}_\text{ENH}$ by its most important components, from low to high frequency components. By setting the $m^\prime$ in Eq.~\ref{svd} to $m^\prime/n^\prime$ where $n^\prime$$>$$1$, we can make a balance between dimensionality reduction and quality of reconstruction. This operation actually acts as principal component analysis \cite{wall2003singular}. Empirically, the magnitudes of $G$ will be less than the pixel precision ($1/255$ for an 8-bit representation) at indices around $n^\prime$=$2$ and therefore they can be pruned without any visual impact on the spectrograms. Though this dimension reduction resizes spectrogram dimension to half, the quality of the reconstructed image is quite good, and differences are imperceptible to the human visual system (see Fig.~\ref{color_space_recon}). The outputs of the dimensionality reduction block in Fig.~\ref{PreprocDiag} are linear, logarithmic, and logarithmic real spectrograms visualized in three color spaces (BBG, PG, and WB) which are all reduced to half of their original dimension. 

\begin{figure}[htpb!]
  \centering
  \includegraphics[width=.35\textwidth]{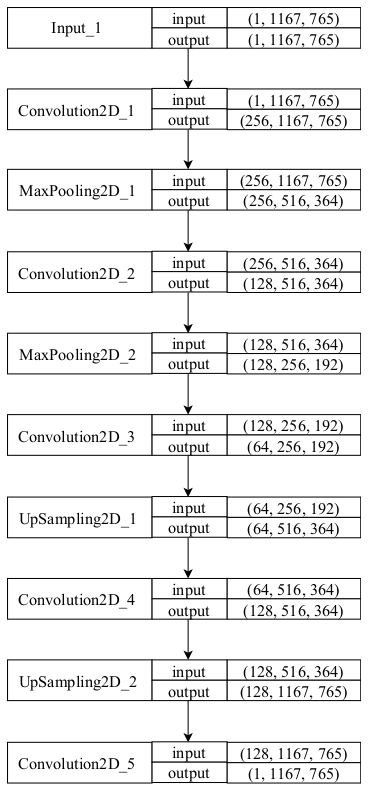}
  \caption{Architecture of the proposed CDA to smooth the spectrogram representations. Numbers inside parentheses denote the number of filters, and width and height of inputs and outputs, respectively.} 
  \label{CDA}
\end{figure}

Though highboost filtering enhances high frequency components in spectrograms and therefore it leads to a better feature extraction, it may also boost noise, especially for the PG and WB color compensated representations. This problem can be minimized to some extent by the dimensionality reduction by SVD, but it is still necessary to improve the quality of the final compensated representations of spectrograms. For addressing this issue, highboost filtered spectrograms are smoothed using a denoising autoencoder with three convolution layers \cite{goodfellow2016deep}. The main advantage of convolutional denoising autoencoder (CDA) over traditional smoothing algorithms is its flexibility in data adaptation and fine reconstruction. Besides, another important reason for using the CDA is to make spectrograms more robust against small adversarial perturbations which machine learning models are very sensitive to. The architecture of the proposed CDA depicted in Fig.~\ref{CDA} is data-dependent and it considers spectrograms of dimension 1167$\times$765 as input. The architecture of the encoder shown in Fig.~\ref{CDA} has three convolutional layers with 5$\times$5 receptive fields, stride 1, {\it Relu} activation function, dropout of 0.5, and two max pooling layers. For corrupting the input data, we used the spectrograms derived from SVD as well as the technique introduced by Vincent \textit{et al.} \cite{vincent2008extracting}. 

Finally, after all these preprocessing steps, the enhanced spectrograms are ready to undergo to feature extraction and classification, as described in the following subsection. Besides that, the enhanced spectrograms can also be used with pre-trained CNN architectures such as AlexNet or GoogLeNet, as described in Section \ref{sec:exp}. 

\subsection{Feature Extraction and Classification}
The proposed approach includes five steps for feature extraction and classification as depicted in Fig.~\ref{Flowchart_Classifier}. The main idea is to extract features from a static sized moving aperture (a.k.a. grid shifting block) which spans a spectrogram with a dynamic stride within a block with dynamic size. Next, we maximize the geometrical distance among feature vectors of different classes and finally we train an SVM classifier on such an organized feature space. Since the proposed approach aims of achieving both classification accuracy and robustness against adversarial attacks, we have evaluated several handcrafted features and representation learning methods to finally come up with SURF instead of CNN features. Empirically, such a feature encoding outperforms DNN features (with/without convolution layers) both in terms of classification accuracy and robustness against adversarial attacks. Our main hypothesis relies on the nature of these features which are projected gradients compared to features generated by DNNs, which generally lead to high classification accuracy but empirically, they have a negative effect on the robustness of the trained model, which becomes quite vulnerable to adversarial attacks. 

\begin{figure}[htpb!]
  \centering
  \includegraphics[width=0.45\textwidth]{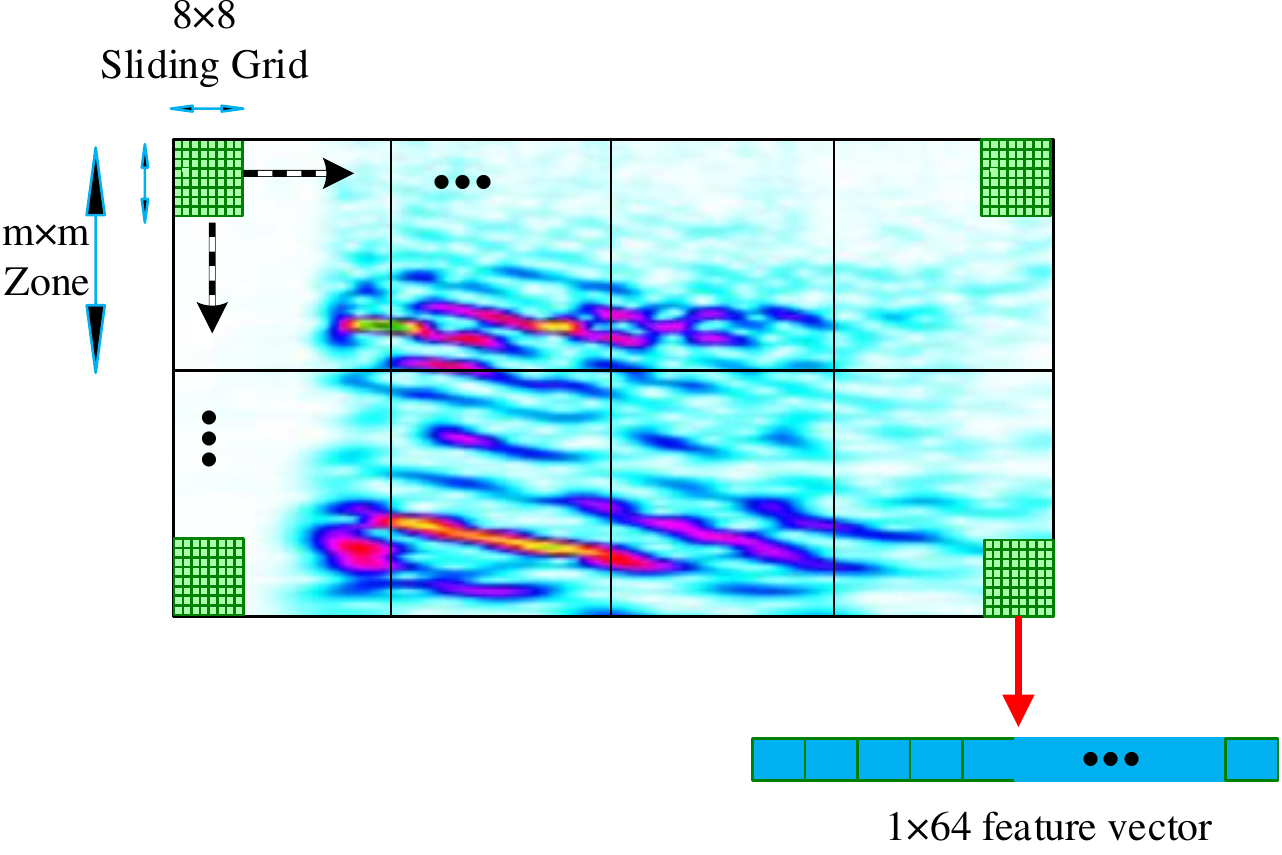}
  \caption{Example of a grid sliding over a spectrogram. The grid slides through the square zone with different strides.}
  \label{Grid_shifting}
\end{figure}

The first step is zoning, which divides a given spectrogram into zones that may vary from 16$\times$16 to 128$\times$128 pixels. Empirically, a zone size of 16$\times$16 is small enough for capturing subtle pixel density changes and a zone size of 128$\times$128 is preferable for regions with less high frequency components. Then, a sliding grid of size 8$\times$8 will span through them. The stride of the sliding grid varies from five to one according to the zone size, with larger strides on larger zones. This scheme supports the idea of a detailed scanning of spectrograms aiming at extracting more discriminant features. Different values have been evaluated for the stride size and finally it ranges from one to five (see Fig.~\ref{Grid_shifting}). 

Different features could be extracted from each 8$\times$8 grid. We also evaluated scale invariant feature transform (SIFT) as a feature extractor \cite{lowe1999object} but decided to use SURF \cite{bay2006surf} because it is much faster than SIFT in runtime, even if it provides fewer feature vectors compared to SIFT. We applied SURF on sliding grids within each zone as shown in Fig. \ref{Flowchart_Classifier}, and at the end, each spectrogram zone is represented by a 64-dimensional feature vector. For increasing the inter-class geometrical distance among extracted feature vectors, the K-means++ algorithm  \cite{arthur2007k} is used to cluster feature vectors into an organized distribution with respect to their geometrical linear distance. Once centroids are found by the clustering algorithm all feature vectors will be mapped into a distance space according to their centroids. We refer readers to \cite{coates2012learning} for further details. Finally, we train a multiclass SVM classifier with polynomial kernel on the transformed feature vectors. We have also evaluated the SVM with radial basis function (RBF) kernel which could not improve the accuracy. In the following section we evaluate the proposed approach on three datasets and compare the results with other state-of-the-art approaches.

\section{Experimental Results}
\label{sec:exp}
We have carried out several experiments on three benchmarking datasets with the aim of: (i) evaluating the detectability of the current adversarial attacks for 2D audio representations; (ii) assessing the performance of the proposed approach on the enhanced spectrograms and compare it with deep architectures (AlexNet and GoogLeNet) that have been used for audio classification; (iii) evaluating the resiliency of the proposed approach and the two deep architectures against several types of adversarial attacks; (iv) characterizing the transferability of the adversarial audio attacks across two different classification paradigms, CNNs and SVMs. 

The UrbanSound8K dataset has 8,732 audio samples of up to four seconds of 10 classes (air conditioner, car horn, children playing, dog bark, drilling, engine idling, gun shot, jackhammer, siren, and street music). The ESC-50 dataset includes 2,000 5-second samples of 50 classes including major groups of animals, natural sound capes \& water sounds, human non-speech sounds, domestic sounds, and exterior noises. The ESC-10 dataset is a subset of ESC-50 which includes 400 recordings of 10 classes (dog bark, rain, sea waves, baby cry, clock tick, person sneeze, helicopter, chainsaw, rooster, and fire crackling). 

\subsection{Detectability of Adversarial Audio Attacks}
\label{sec:detaudio}
The definition of an adversarial attack relies on whether the attack is easily identified or not. We have carried out some experiments to evaluate two of the most powerful adversarial attacks on audio: Backdoor and the DolphinAttack. For such an aim, we generated short-time Fourier transform (STFT), DWT, and cross recurrence plot (CRP) spectrograms for the audio samples of the UrbanSound8K dataset and computed the LID score considering different values of $k$ as shown in Eq.~\ref{LID}. Basically, the LID score should be able to discriminate between negative and positive classes which means returning higher values. In other words, small values of the LID score denote an indistinguishable difference between positive and negative classes which can in turn be interpreted as positive classes may not be considered as adversarial. Table~\ref{LID_Table} shows that the differences between LID scores of positive and negative classes are quite small and it also shows the very low accuracy of the logistic regression classifier trained on these classes. As Table~\ref{LID_Table} shows, legitimate, noisy, and adversarial examples lie in the same subspace and in fact they lie in the same manifolds because they have very similar LID scores. In other words, the adversarial examples generated by both Backdoor and DolphinAttack are almost equivalent to examples corrupted by random noise, which basically does not seem to satisfy the definition of adversarial examples. Moreover, the performance of the logistic regression is quite low and shows poor discrimination between negative and positive classes which should be higher than 60\%. 

\begin{table}[htpb!]
\centering
\renewcommand{\arraystretch}{1.1}
\footnotesize
\caption{LID score for different representations of UrbanSound8K samples. Mean difference is generated for two classes of negative (legitimate and random noisy) and positive (adversarial by Backdoor and DolphinAttack).} 
\begin{tabular}{|c||c||c||c|}
\hline
\multirow{2}{*}{Representation} &  \multirow{2}{*}{$k$} & Mean Difference & Classification \\   
&& of LID Scores & Accuracy (\%) \\ \hline 
\multirow{4}{*}{DWT} & 50  & 0.082& 11.23\\ \cline{2-4} 
  & 75  & 0.071& 10.04\\ \cline{2-4} 
  & 100 & 0.036& 10.01\\ \cline{2-4} 
  & 125 & 0.032& 09.46\\ \hline
\multirow{4}{*}{STFT} & 50  & 0.076& 13.05\\ \cline{2-4} 
  & 75  & 0.074& 12.94\\ \cline{2-4} 
  & 100 & 0.066& 12.92\\ \cline{2-4} 
  & 125 & 0.061& 11.87\\ \hline
\multirow{4}{*}{CRP}  & 50  & 0.089& 15.01\\ \cline{2-4} 
  & 75  & 0.084& 14.56\\ \cline{2-4} 
  & 100 & 0.079& 14.32\\ \cline{2-4} 
  & 125 & 0.078& 13.77\\ \hline
\end{tabular}
\label{LID_Table}
\end{table}

\subsection{Accuracy and Resilience of CNNs and SVMs}
Deep neural networks require a large amount of data for training. For increasing the size of datasets aiming at extracting more information from them, we augmented the number of samples by stretching (speeding up) and shrinking (slowing down) recordings in time (pitch shifting) using MUDA library \cite{mcfee2015software}. This is a common approach in sound processing which affects favourably the classifier's performance \cite{salamon2017deep}. The scale values that were applied for pitch-shifting are: 0.5, 0.75, 0.9, 1.1, 1.25, 1.5, and 1.75. This operation increases the size of each dataset in eight times. 

\begin{table}[ht!]
\footnotesize
\centering
\renewcommand{\arraystretch}{1.1}
\caption{Scale operators ($c$) for color compensation.}
\label{scale_val}
\begin{tabular}{|c||c||c|}
\hline
Dataset   & \multicolumn{1}{l||}{Color Compensation} & $c$ \\ \hline
\multirow{3}{*}{ESC-10}  & BBG&0.57\\ \cline{2-3} 
   & PG &  0.74  \\ \cline{2-3} 
   & WB &  0.46  \\ \hline
\multirow{3}{*}{ESC-50}  & BBG&0.81\\ \cline{2-3} 
   & PG &   0.79 \\ \cline{2-3} 
   & WB &  0.58  \\ \hline
\multirow{3}{*}{UrbS8K} & BBG&  0.72  \\ \cline{2-3} 
   & PG & 0.85   \\ \cline{2-3} 
   & WB &  0.67  \\ \hline
\end{tabular}
\end{table}

For generating the spectrogram $\bold{sp}_\text{STFT}$, we used the approach suggested by Boddapati \textit{et al.}~\cite{boddapati2017classifying} by setting sampling frequency to 8 kHz, 16 kHz, and 8 kHz for ESC-10, ESC-50, and UrbanSound8K datasets, respectively. Also, the frame length was set to 50 ms (ESC-10), 30 ms (ESC-50), and 50 ms (UrbanSound8K) with a fixed overlapping of 50\%. These values have been found after conducting exploratory experiments on these datasets. For generating the spectrogram $\bold{sp}_\text{DWT}$, we used 256 frequency bins with a Morlet mother function as proposed by Cowling and Sitte \cite{cowling2003comparison} and linear, logarithmic, and logarithmic real magnitude scales for enhancing high, low and medium frequencies, respectively. The scale operators $c$ as described in Eq.~\ref{highboost}, are shown in Table~\ref{scale_val}. The SVM uses a quadratic kernel with the cost parameter $\left \| c \right \|\leq 0.1$ and the kernel parameter $\left \|  \gamma \right \| < 0.003$. Besides the quadratic kernel, we also evaluated a linear SVM, which is referred simply as SVM in several tables in this section. We have used the scikit-learn \cite{pedregosa2011scikit} package for implementing SVMs. 

In the first experiment, we trained AlexNet and GoogLeNet with the same setup proposed by Boddapati \textit{et al.}~\cite{boddapati2017classifying} which leads to the highest classification performance reported in the literature for 2D representations. These two deep convolutional neural networks were trained on a linear pooling of STFT ($\bold{sp}_\text{STFT}$), MFCC ($\bold{sp}_{\text{MFCC}}$), and CRP ($\bold{sp}_{\text{CRP}}$) spectrograms, as:
\begin{equation}
\bold{sp}_\text{POOL} = \mathrm{clip}\begin{pmatrix}
\bold{sp}_{\text{STFT}}+\bold{sp}_{\text{MFCC}}+\bold{sp}_{\text{CRP}} , [0, 1]
\end{pmatrix}
\label{frmfcccrp}
\end{equation}
\noindent where $\bold{sp}_\text{POOL}$ denotes the resulting pooled spectrogram which values outside the range [0,1] are clipped to the value at the boundary of the range. These spectrograms are computed for the three environmental sound datasets (ESC-10, ESC-50, and UrbanSound8K) after the augmentation procedure. In addition to training our classifier on the pooled representation, referred to as POOL, we also trained it on the enhanced 2D representation space as shown in Fig.~\ref{PreprocDiag}, referred to as DWT. These two representations are also evaluated for AlexNet and GoogLeNet. In other words, we evaluate the performance of AlexNet and GoogLeNet on the spectrograms obtained from our data preprocessing approach. These two experiments are executed using 5-fold cross validation with a ratio of 0.2 for testing. We used four parallel GPUs GTX580 based on an implementation based on \cite{krizhevsky2012imagenet}. We stopped training after 83 epochs using early stopping for AlexNet and GoogLeNet. The results achieved by these two classifiers are reported in Table~\ref{Rec_acc}. As Table~\ref{Rec_acc} shows, AlexNet and GoogLeNet have achieved the best performances for both representation spaces, although the proposed approach presents competitive results. The differences between the best deep model and the proposed approach range from 4.32\% for UrbanSound8K to 11.02\% for ESC-50. We also repeated this experiment with 10-fold cross validation as suggested in \cite{salamon2017deep}, but the results were very close to those reported in Table~\ref{Rec_acc}. 


\begin{table}[htpb!]
\footnotesize
\centering
\renewcommand{\arraystretch}{1.1}
\caption{Mean classification accuracy (5-fold CV) of four classifiers on two representation spaces: POOL and DWT. The best performances are shown in bold.}
\label{Rec_acc}
\begin{tabular}{|c||c||c||c||c||c|}
\hline
\multirow{2}{*}{Dataset}   &  \multirow{2}{*}{Repres.} & \multicolumn{4}{c|}{Mean Accuracy (\%)} \\ \cline{3-6}
 & & \multicolumn{1}{l||}{GoogLeNet} & \multicolumn{1}{l||}{AlexNet} & \multicolumn{1}{l||}{SVM} & \multicolumn{1}{l|}{Proposed}  \\ \hline
\multirow{2}{*}{ESC-10}  & POOL & \textbf{83.19}  &   82.54 & 64.23 & 78.31 \\ \cline{2-6} 
   & DWT   &  \textbf{83.21}  &  82.90 & 70.45 & 79.10\\ \hline
\multirow{2}{*}{ESC-50}  & POOL & \textbf{71.36}  & 64.09 & 52.37 & 60.10\\ \cline{2-6} 
   & DWT&  \textbf{71.20}  & 66.41 & 55.09 & 60.41\\ \hline
\multirow{2}{*}{UrbS8K} & POOL & \textbf{91.08} & 90.06 & 72.03 & 86.15\\ \cline{2-6} 
   & DWT& 86.85 & \textbf{90.10}  & 72.89 & 86.39\\ \hline
\end{tabular}
\end{table}

However, a high accuracy does not translate to a high robustness against adversarial attacks. In Table~\ref{Transferability}, we assess the robustness of the classifiers of Table~\ref{Rec_acc} against several adversarial attacks as well as the transferability of such adversarial attacks across different models. For such an aim we have developed the FGSM, BIM-a, BIM-b, and CWA adversarial attacks (deep model attacks) for AlexNet and GoogLeNet and the EA and regular Evasion attacks (SVM attacks) for SVM classifiers. The total number of adversarial examples crafted using each attack for different datasets is equivalent to the number of samples in the legitimate dataset. In other words, for each legitimate sample, one adversarial example is crafted by each adversarial attack algorithm. Since FGSM and CWA are targeted, adversarial examples of these two attacks are crafted toward a random wrong label. This not only makes our evaluations fair against non-targeted attacks, but also reduces the cost of crafting adversarial examples of datasets with more than 10 classes, which is the case of the ESC-50 dataset that has 50 classes. Then, these crafted examples are fed to both deep learning and SVM models to compute the ratio of successful fooling over the total number of adversarial examples (fooling rate) in a black-box scenario. 

\begin{table}[htpb!]
\centering
\footnotesize
\renewcommand{\arraystretch}{1.1}
\caption{Mean fooling rate (5-fold CV) of two CNNs and two SVMs against six strong adversarial attacks. The best performances are shown in bold (lowest values).}
\label{Transferability}
\begin{tabular}{|c||c||c||c||c||c|}
\hline
 Dataset & Adv. & \multicolumn{4}{c|}{Mean Fooling Rate (\%)}  \\ \cline{3-6}
(Repres.)& Attack & GoogLeNet & AlexNet & SVM & Proposed \\ \hline
\multirow{4}{*}{\begin{tabular}[c]{@{}c@{}}ESC-10\\ (POOL)\end{tabular}}  & FGSM   & 95.23&  94.04  &   60.78&\textbf{43.12} \\ \cline{2-6} 
& BIM-a  &94.07 &   90.13 &   61.68& \textbf{48.60}\\ \cline{2-6} 
& BIM-b  &94.26 &91.30&62.46   &\textbf{46.03}\\ \cline{2-6}
& CWA&95.89 &   93.66 &94.01 &   \textbf{51.77}\\ \cline{2-6}  
& LFA  &\textbf{51.23} &63.01&94.43   & 60.47\\ \cline{2-6}& EA&\textbf{43.79} &   44.12 &94.14   &   58.34\\ \hline
\multirow{4}{*}{\begin{tabular}[c]{@{}c@{}}ESC-10\\ (DWT)\end{tabular}}& FGSM   &94.30 &   93.36 &   64.05&   \textbf{50.02}   \\ \cline{2-6} 
& BIM-a  & 92.15&92.87& 59.57  & \textbf{51.13}\\ \cline{2-6} 
& BIM-b  &93.58 &   92.33 &57.92   &\textbf{43.07} \\ \cline{2-6} 
& CWA&95.36 &   94.89 &64.35 &   \textbf{53.18}\\ \cline{2-6}  

& LFA  &57.36 &\textbf{56.35}&95.58   & 71.64\\ \cline{2-6}  
& EA&49.66 &   \textbf{48.00} &92.89&  61.78\\ \hline
\multirow{4}{*}{\begin{tabular}[c]{@{}c@{}}ESC-50\\ (POOL)\end{tabular}}  & FGSM   &96.78 &95.61&69.22&\textbf{51.99}  \\ \cline{2-6} 
& BIM-a  & 95.01&   96.08 & 67.17  & \textbf{50.20}\\ \cline{2-6} 
& BIM-b  & 94.77&95.17&69.71&\textbf{ 50.03}\\ \cline{2-6} 
& CWA&96.02 &   97.14 &72.10 &   \textbf{53.04}\\ \cline{2-6}  

& LFA  &62.12 &66.35&95.27   & \textbf{60.25}\\ \cline{2-6} & EA& 55.47&\textbf{52.01}& 95.94  &59.03  \\ \hline
\multirow{4}{*}{\begin{tabular}[c]{@{}c@{}}ESC-50\\ (DWT)\end{tabular}}& FGSM   &96.30 &   95.80 & 66.16  &\textbf{50.01}  \\ \cline{2-6} 
& BIM-a  &93.36  &94.05&69.02   &   \textbf{49.36}   \\ \cline{2-6} 
& BIM-b  &91.25  &92.53&67.11   &   \textbf{45.92}  \\ \cline{2-6} 
& CWA&95.73 &   94.11 &70.09 &   \bf 49.31\\ \cline{2-6}  

& LFA  &60.08 &\bf 58.01&92.21   & 62.84\\ \cline{2-6}

& EA&51.37 &   \textbf{49.61} & 90.36  &   58.15   \\ \hline
\multirow{4}{*}{\begin{tabular}[c]{@{}c@{}}UrbS8K\\ (POOL)\end{tabular}} & FGSM   & 94.68&93.22&  60.50 & \textbf{45.17}\\ \cline{2-6} 
& BIM-a  & 94.65&95.32& 58.22  &  \textbf{42.36}\\ \cline{2-6} 
& BIM-b  &90.22 &   91.24 &53.39&  \textbf{42.16}\\ \cline{2-6} 
& CWA&92.08 &   93.62 &\textbf{60.17} &   60.25\\ \cline{2-6}  

& LFA  &\textbf{ 55.01} &78.36&96.14   & 65.35\\ \cline{2-6}

& EA& 44.02&   \textbf{41.07}& 95.16  &62.30  \\ \hline
\multirow{4}{*}{\begin{tabular}[c]{@{}c@{}}UrbS8K\\ (DWT)\end{tabular}}& FGSM   &94.14 &93.02&  57.31 &\textbf{48.33}  \\ \cline{2-6} 
& BIM-a  & 92.43&93.21&62.01   & \textbf{51.07}\\ \cline{2-6} 
& BIM-b  &94.01 &   93.61 &63.32&  \textbf{53.03}\\ \cline{2-6} 
& CWA&95.27 &   93.84 &62.14 &   \textbf{50.48}\\ \cline{2-6}  
  
& LFA  &\textbf{ 54.33} &55.03&92.06   & 63.52 \\ \cline{2-6}  
& EA& 47.01 &\textbf{45.50}& 91.02  &  59.01\\ \hline
\end{tabular}
\end{table}

 Table~\ref{Transferability} also shows the transferability of adversarial attacks crafted to attack deep models to attack SVM models and vice-versa. A high adversarial transferability rate represents a serious threat for data-driven classifiers. In other words, a reliable classifier should not only be robust against adversarial attacks designed to fool its own type of model, but it should also be reasonably resistant against attacks designed to attack other types of model. Table~\ref{Transferability} shows the results achieved on both experiments. The mean fooling rate, which measures the success rate of adversarial examples in fooling the machine learning models in terms of the percentage of adversarial samples misclassified by the models is computed for comparing the performance of CNNs and SVMs against the six adversarial attacks. Table~\ref{Transferability} shows that both for both CNNs and SVMs are quite vulnerable to the adversarial attacks designed to attack its own model, with fooling rates higher than 90\%. However, the proposed approach not only is quite robust, but also has the lowest fooling rate against adversarial attacks (EA and LFA) designed for such a model, with fooling rates between 58.15\% and 71.64\%. Table~\ref{Transferability} also reveals that, there is a higher chance of fooling SVM models by deep attacks compared to fooling AlexNet and GoogLeNet by adversarial examples crafted by EA or LFA. Additionally, AlexNet is more robust against SVM-based adversarial attacks compared to GoogLeNet, though its mean accuracy is a little lower than GoogLeNet. 

\begin{table}[htpb!]
\footnotesize
\renewcommand{\arraystretch}{1.1}
\centering
\caption{Average ranking considering the mean accuracy and the fooling rate for all models, datasets and adversarial attacks.}
\begin{tabular}{|c||c||c||c||c|}
\hline
\multirow{2}{*}{Approach} & \multicolumn{2}{c||}{Mean Accuracy} & \multicolumn{2}{c|}{Fooling Rate} \\
 \cline{2-5}
 & $\bar{r}$ & Rank & $\bar{r}$ & Rank \\
\hline
GoogLeNet & 1.17 & 1 & 2.97 & 4\\
\hline
AlexNet & 1.83 & 2 & 2.78 & 3\\
\hline
SVM & 4.00 & 4 & 2.67 & 2 \\
\hline
Proposed  & 3.00 & 3 & 1.61 & 1 \\
\hline
\end{tabular}
\label{tab:avgrank}
\end{table}

\begin{figure}[htpb!]
  \centering
  \includegraphics[width=0.48\textwidth]{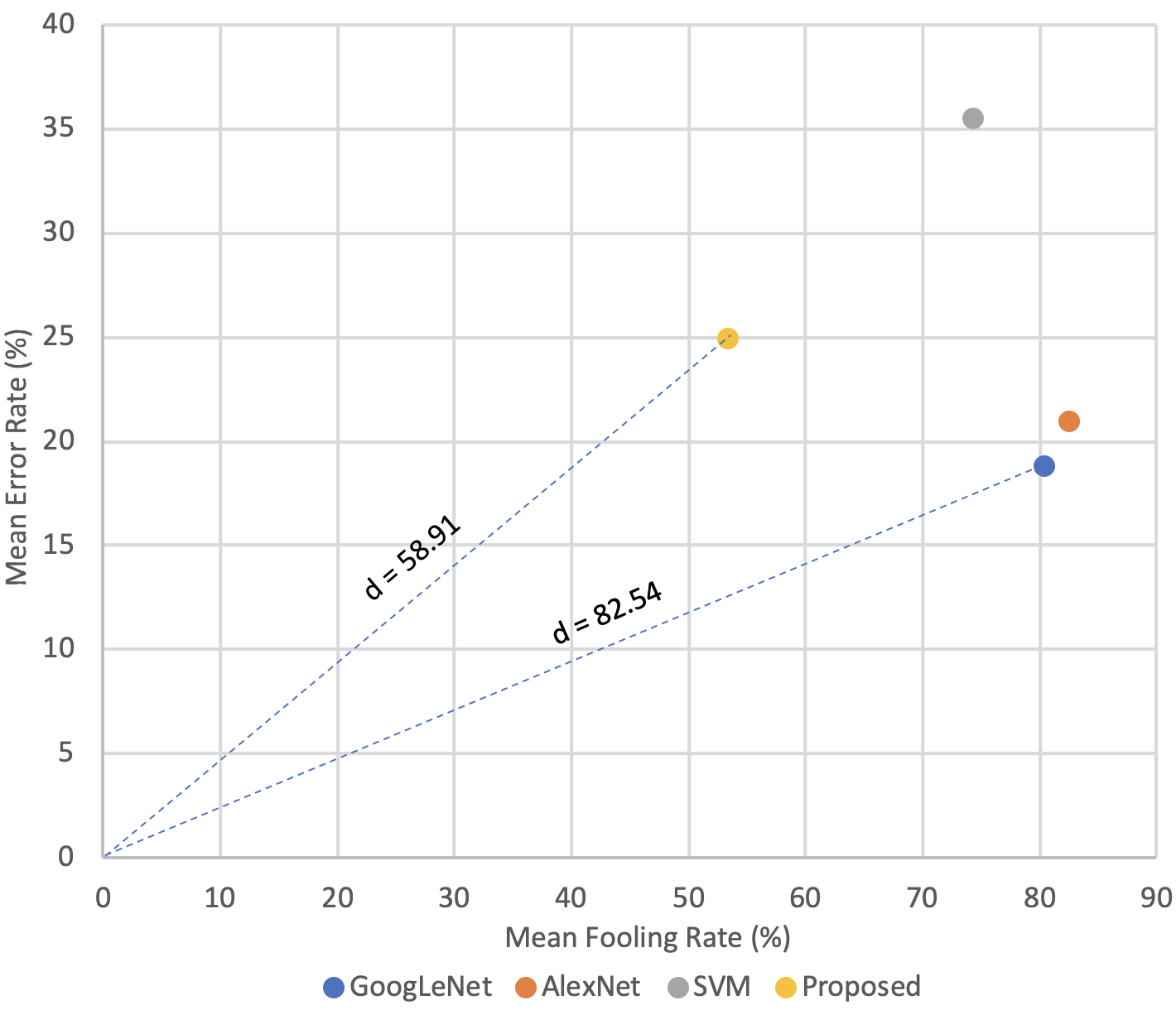}
  \caption{Comparison between deep models, SVM, and the proposed approach over all datasets and representations of Table~\ref{Rec_acc}. The Euclidean distance to the origin for the two best approaches is also shown. }
  \label{fig:errorXfooling}
\end{figure}

Table~\ref{tab:avgrank} shows average rankings of our evaluation metrics of recognition accuracy and fooling rate with respect to the statistics provided in Table~\ref{Transferability}. Regarding this table, the smaller the $\bar{r}$ is, the better are the accuracy and the fooling rates. Although the proposed approach appears in third place in the mean accuracy rank, it is the first one in resiliency against the six types of adversarial attacks. Therefore, this indicates a good tradeoff between accuracy and resiliency. This is also shown in Fig.~\ref{fig:errorXfooling}, where the proposed approach is the one closest to the origin (zero error rate and zero fooling rate) according to the Euclidean distance ($d$ = 58.91). Fig.~\ref{fig:errorXfooling} also shows that while the mean error rate of the proposed approach is 6.07\% higher than GoogLeNet, the proposed approach is 26.96\% more robust to adversarial attacks than GoogLeNet. Furthermore, the mean error rate of the proposed approach is 10.57\% lower than the SVM and it is also 20.98\% more robust to adversarial attacks. Notwithstanding the good tradeoff achieved by the proposed approach, there is still a large room for improvements.
\subsection{Analysis of the Proposed Approach}
The proposed approach provides the best tradeoff between accuracy and resilience to adversarial attacks than deep models and SVM according to the proposed metric shown in Fig.~\ref{fig:errorXfooling}. For understanding the reason(s) of such a best tradeoff, we dig into the preprocessing (Fig.~\ref{PreprocDiag}) of the proposed approach. We safely remove each module (or submodule) from the preprocessing part and measure its positive or negative contribution to the mean accuracy and robustness against the six types of adversarial attacks. Table~\ref{tab:remove_module} reveals that the proposed approach takes advantage of both CDA and SVD compression. The most straightforward impact of these two operations is in affecting (smoothing) high frequency components where subtle changes of adversarial examples probably lie on. It has been proved that autoencoders can clean adversarial examples and therefore defend the targeted trained models from the adversarial attacks \cite{nayebi2017biologically, meng2017magnet}. Moreover, for measuring the effect of the first two modules of Fig.~\ref{Flowchart_Classifier} on final classification performance, we carried out some additional experiments including removing them and changing block size and grid shifting stride on a 5-fold cross validation. In Table~\ref{zone_grid}, we only report some of the highest mean accuracy with respect to zoning size and grid shifting stride.  

\begin{table}[htpb!]
\centering
\footnotesize
\renewcommand{\arraystretch}{1.1}
\caption{The average effect of removing each module from Fig.~\ref{PreprocDiag} on the mean accuracy and robustness of the proposed model against deep and SVM adversarial attacks. Positive ($+$) and negative ($-$) effects are shown by their signs.}
\label{tab:remove_module}
\begin{tabular}{|c||c||c||c|}
\hline
\multirow{3}{*}{Module} & Mean & \multicolumn{2}{c|}{Robustness Against (\%)} \\ \cline{3-4}
& Accuracy (\%) & SVM Attacks & Deep Attacks \\ 
\hline
Spectr. Vis. & $-$16.47 & $-$4.07 & $-$3.14 \\ \hline
Color Comp. & $-$7.36&$-$0.36   &$+$2.64\\ \hline
Highb. Filt. &  $-$9.52  &$-$0.75   &$-$1.96 \\ \hline
SVD   & $-$8.21&$-$6.18&$-$4.17\\ \hline
CDA   &  $-$7.94  &$-$9.18   &$-$6.33\\ \hline
\end{tabular}
\end{table}

\begin{table}[htpb!]
\footnotesize
\centering
\renewcommand{\arraystretch}{1.1}
\caption{The effect of selected zoning size and shifting grid length on the overall recognition accuracy of the proposed approach on DWT representation of the UrbanSound8K dataset.}
\label{zone_grid}
\begin{tabular}{|c||c||c|}
\hline
Zoning Size & Sliding Grid Stride & Mean Accuracy (\%)\\
\hline
{[}16, 32, 64, 96, 128{]} & {[}1, 2, 3, 4, 5{]}  & \bfseries 79.33\\ \hline
{[}16, 32, 64, 96, 128{]} & {[}2, 2, 2, 2, 2{]}  & 77.29\\ \hline
{[}8, 16, 32, 64, 128{]}  & {[}1, 2, 3, 4, 5{]}  & 76.18\\ \hline
{[}16, 32, 64, 96, 128{]} & {[}4, 3, 3, 3, 4{]}  & 74.22\\ \hline
{[}32, 64, 128{]}& {[}3, 2, 1{]}   & 73.91\\ \hline
{[}64, 96, 128{]}& {[}3, 2, 1{]}   & 72.63\\ \hline
None& None   & 70.92\\ \hline
\end{tabular}
\end{table}

\section{Conclusion}
In this paper, we discussed the serious threat that adversarial attacks may pose to machine learning models trained either on 1D or 2D audio representations. While there is no reliable adversarial attack on raw audio signals, there is a bijective relation between 1D signals and spectrograms which opens the avenue for adversarial transferability between these two representation spaces and that poses a real security concern. Besides that, considering that the majority of state-of-the-art approaches for audio classification rely on 2D representations, most of them based on CNNs originally designed for image classification tasks, we showed that CNNs trained on spectrograms of environmental sound signals achieve state-of-the-art performance in terms of accuracy. However, these CNNs are not reliable at all, as they can be easily fooled by adversarial examples, with fooling rates higher than 90\%. 

Therefore, we proposed a novel approach for environmental sound classification based on 2D representations that provides a good tradeoff between accuracy and resiliency to the most powerful adversarial attacks designed to fool both deep neural models and SVMs. The proposed approach was compared to AlexNet, GoogLeNet, and a linear SVM classifier on three publicly available datasets. The highest mean recognition rates were achieved by GoogLeNet (81.15\%), AlexNet (79.15\%), the proposed approach (75.08\%), and the linear SVM (64.51\%), respectively. However, in addition to the competitive accuracy, the proposed approach outperforms by far all three mentioned classifiers in terms of robustness against adversarial attacks since the mean fooling rates for these four models are 95.15\%, 94.36\%, 50.56\%, and 66.74\% considering deep attacks and 52.62\%, 54.79\%, 61.89\%, and 93.77\% considering SVM attacks. However, as shown in Fig.~\ref{fig:errorXfooling}, there is still a large room for improvements. As a future study, we are interested in employing reactive adversarial detection algorithms (e.g., LID detector) as a postprocessing operation aiming at increasing the robustness of the proposed approach. 

We are also inclined to explore the resiliency of our classification scheme for raw audio signals rather than spectrograms against audio attacks and measure its capability against audio played back over the air. To this end, we may need to remove/add some preprocessing steps which have shown positive impacts on the robustness of the proposed approach against adversarial attacks (e.g. CDA); and consequently, simplify our approach which requires several steps of processing. Another important aspect that deserves further studies is the adversarial example transferability bijectively from 1D audio signal to 2D spectrograms and vice versa. In other words, we would like to explore the possibility of crafting adversarial audio examples for a model trained on 1D signals and transfer such an attack to the 2D representation to be able to fool a 2D model trained on spectrograms, and also the other way around. Since many audio classification approaches implement different types (ensemble) of data-driven models (both 1D and 2D) aiming at improving their prediction confidence, hence if a crafted adversarial example can fool both 1D and 2D models, it may constitute a true threat to several sound recognition/processing systems and devices (e.g. voice id devices). 

\section*{Acknowledgment}
This work was funded by the Natural Sciences and Engineering Research Council of Canada (NSERC) under Grant RGPIN 2016-04855 and Grant RGPIN 2016-06628.

\ifCLASSOPTIONcaptionsoff
  \newpage
\fi



\bibliographystyle{IEEEtran}
\bibliography{IEEEabrv,mybib}
%
%
\newpage
\begin{IEEEbiography}[{\includegraphics[width=1.0in,height=1.25in,clip]{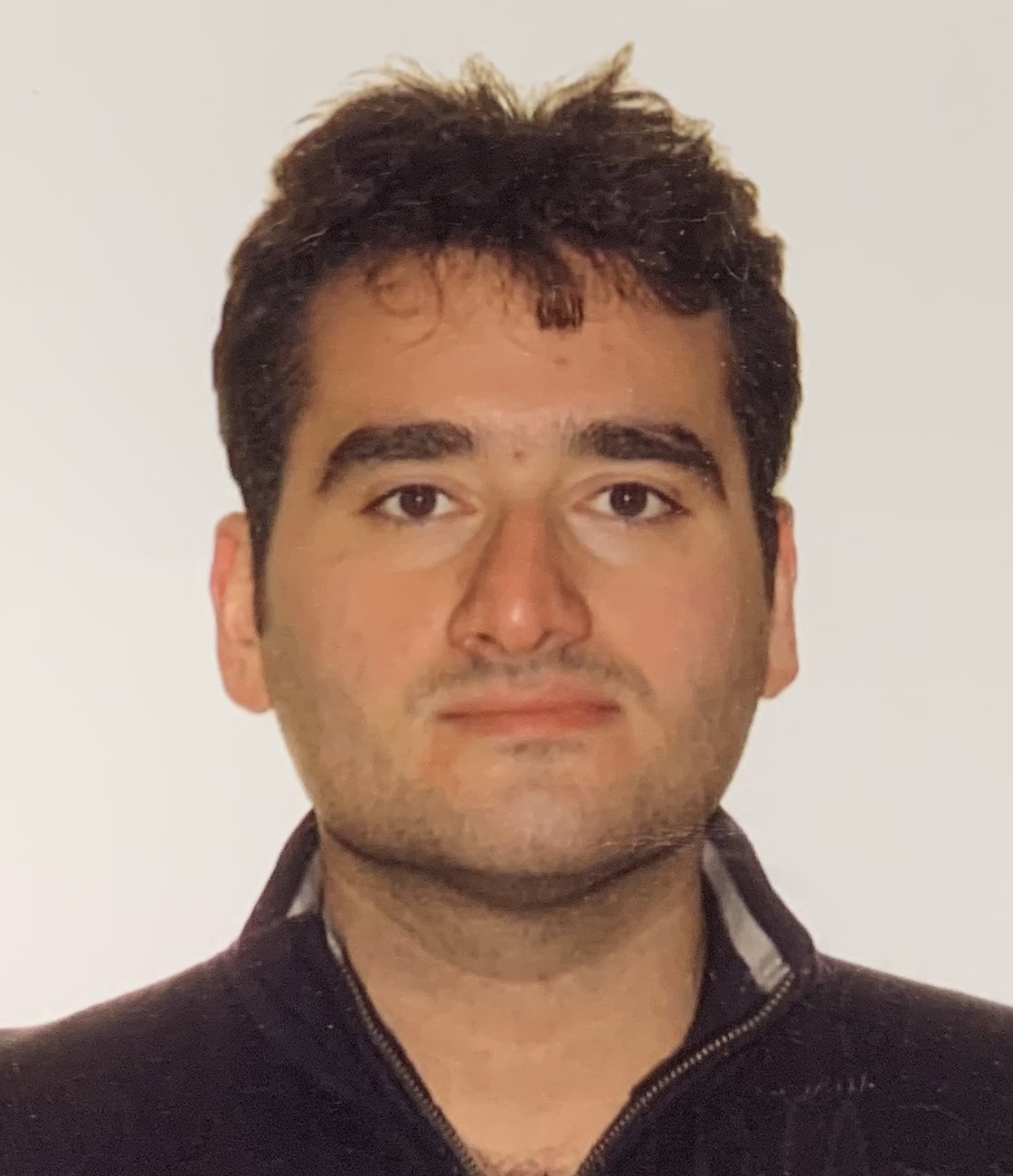}}]
{Mohammad Esmaeilpour}
received his B.Sc. degree in computer software engineering in 2012, M.Sc. of artificial intelligence in 2014 (recognized as top straight A+ student), and currently doing his Ph.D. at \'{E}cole de Technologie Sup\'{e}rieure (\'{E}TS), Montr\'{e}al, Qu\'{e}bec. As an experienced physically-based animation programmer, he has been involved in intern character locomotion modeling projects at Ubisoft gaming company. In an industrial collaboration with AudioZ company in Montr\'{e}al, he serves as a lead programmer in voice activity/event detection project for speech processing systems. His research interests include developing machine learning algorithms for high dimensional data such as animation, wild audio, and speech.
\end{IEEEbiography}
\begin{IEEEbiography}
[{\includegraphics[width=1.0in,height=1.25in,clip]{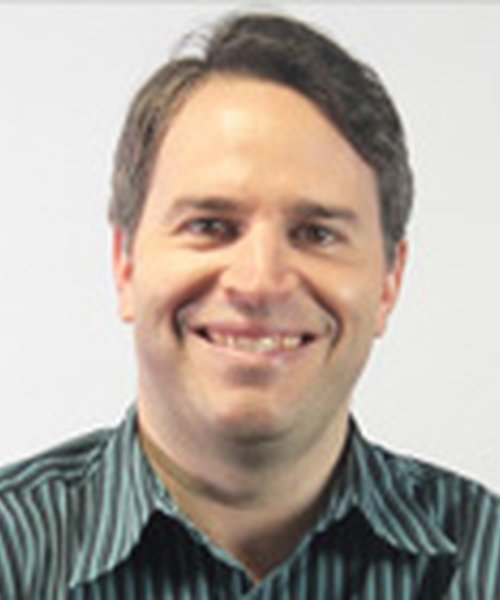}}]
{Patrick Cardinal} received the B. Eng. degree in electrical engineering in 2000 from \'{E}cole de Technologie Sup\'{e}rieure (\'{E}TS), M.Sc. from McGill University in 2003 and PhD from \'{E}TS in 2013. From 2000 to 2013, he has been involved in several projects related to speech processing, especially in the development of a closed-captioning system for live television shows based on automatic speech recognition. After his postdoc at MIT, he joined \'{E}TS as a professor. His research interests cover several aspects of speech processing for real life and medical applications.
\end{IEEEbiography}
\begin{IEEEbiography}
[{\includegraphics[width=1.0in,height=1.25in,clip]{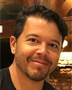}}]
{Alessandro Lameiras Koerich} is an Associate Professor in the Dept. of Software and IT Engineering of the \'{E}cole de Technologie Sup\'{e}rieure (\'{E}TS), University of Qu\'{e}bec, Montr\'{e}al, Canada. He received the B.Sc. degree in electrical engineering from the Federal University of Santa Catarina, Brazil, in 1995, the MSc degree in electrical engineering from the University of Campinas, Brazil, in 1997, and the Ph.D. degree in engineering from the \'{E}TS, in 2002. From 1997 to 1998, he was a lecturer at the Federal Technological University of Paran\'{a}. From 1998 to 2002, he was a visiting scientist at the CENPARMI, Montr\'{e}al, Canada. From 2003 to 2015 he was with the Pontifical Catholic University of Parana, Curitiba, Brazil, where he became professor in 2010 and served as chair of Graduate Studies in CS from 2006 to 2008. From 2009 to 2015 he was also an associate professor in the Dept. of Electrical Engineering of Federal University of Paran\'{a}. In 2004, he was nominated IEEE CS Latin America Distinguished Speaker. He was a visiting researcher at INESC-Porto, Portugal from 2009 to 2012 and served as a Fulbright Visiting Professor in the Dept. of Electrical Engineering at Columbia University, New York, USA, in 2013. Prof. Koerich is the author of more than 100 papers and holds four patents. He is an associate editor of the Pattern Recognition journal and served as the general chair of the 14th Intl Society for Music Information Retrieval Conference, which was held in Curitiba, Brazil in 2013. His current research interests include computer vision, machine learning and music information retrieval.
\end{IEEEbiography}

\vfill

\end{document}